\documentclass{article}

\usepackage{microtype}
\usepackage{graphicx}
\usepackage{subcaption}
\usepackage{booktabs} 
\usepackage{multirow}
\usepackage[export]{adjustbox}

\usepackage{hyperref}
\usepackage{todonotes}
\usepackage{amsmath}
\usepackage{amssymb}
\usepackage{nicefrac}
\usepackage[inline]{enumitem}

\DeclareMathOperator*{\argmin}{arg\,min}
\graphicspath{{fig/}}

\usepackage[accepted]{icml2020}

\icmltitlerunning{Contrastive GNN Explanation}

\begin{document}

\twocolumn[
\icmltitle{Contrastive Graph Neural Network Explanation}



\icmlsetsymbol{equal}{*}

\begin{icmlauthorlist}
\icmlauthor{Lukas Faber}{equal,eth}
\icmlauthor{Amin K. Moghaddam}{equal,eth}
\icmlauthor{Roger Wattenhofer}{equal,eth}
\end{icmlauthorlist}

\icmlaffiliation{eth}{ETH Zurich, Switzerland}
\icmlcorrespondingauthor{Lukas Faber}{lfaber@ethz.ch}
\icmlcorrespondingauthor{Amin K. Moghaddam}{khashkhm@ethz.ch}

\icmlkeywords{Machine Learning, ICML, Graph Neural Network, GNN}

\vskip 0.3in
]



\printAffiliationsAndNotice{\icmlEqualContribution} 

\begin{abstract}
Graph Neural Networks achieve remarkable results on problems with structured data but come as black-box predictors.
Transferring existing explanation techniques, such as occlusion, fails as even removing a single node or edge can lead to drastic changes in the graph.
The resulting graphs can differ from all training examples, causing model confusion and wrong explanations.
Thus, we argue that explicability must use graphs compliant with the distribution underlying the training data.
We coin this property \emph{Distribution Compliant Explanation} (DCE) and present a novel Contrastive GNN Explanation (CoGE) technique following this paradigm.
An experimental study supports the efficacy of CoGE.
\end{abstract}

\section{Introduction}
\label{intro}
While neural networks have shown many breakthroughs, they come as black-box predictors. Thus, research has developed explicability techniques to shed light on how a neural network makes decisions. Such explanations help to understand and trust the model. One approach is to identify those parts of the input, which are most influential in generating the output. With Graph Neural Networks (GNNs) being generalizations of Convolutional Neural Networks (CNNs), one might hope that the techniques for image explicability of CNNs transfer to GNNs. 

For example, we could apply occlusion~\citep{zeiler2014visualizing, ancona2017unified} to GNNs. In computer vision, occlusion measures the importance of a pixel by removing the pixel from the image, e.g., by setting it black. Similarly, we could measure the importance of a node or edge by removing this node or edge. However, such removals are drastic. Especially in sparse graphs node or edge removals may change the topology of the graph completely.

\begin{figure}[ht]
\vskip 0.2in
\begin{center}
\begin{subfigure}{0.23\textwidth}
\includegraphics[width=0.8\linewidth]{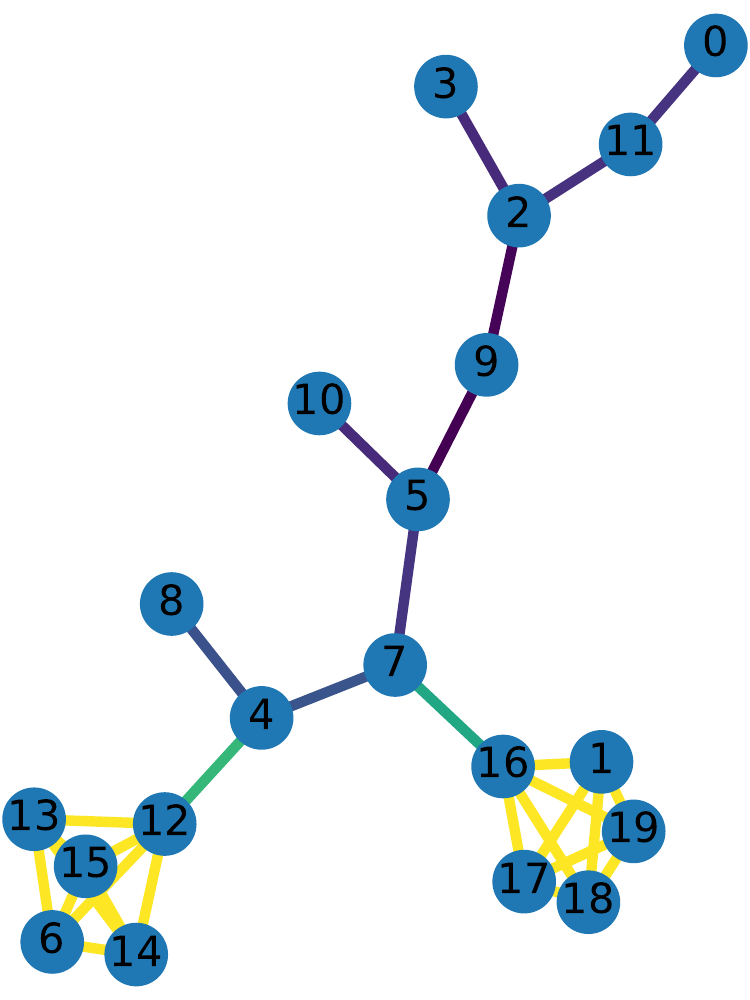} 
\caption{CoGE}
\label{fig_ours}
\end{subfigure}%
\begin{subfigure}{0.23\textwidth}
\includegraphics[width=0.8\linewidth]{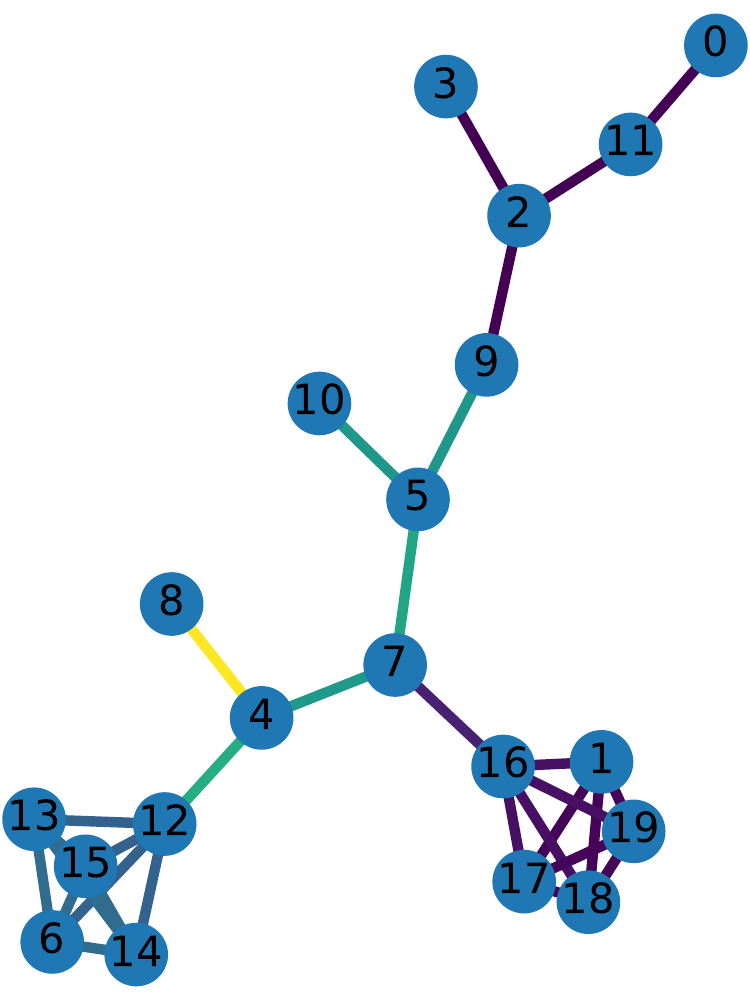}
\caption{Occlusion}
\label{fig_occlusion}
\end{subfigure}
\begin{subfigure}{0.23\textwidth}
\includegraphics[width=0.8\linewidth]{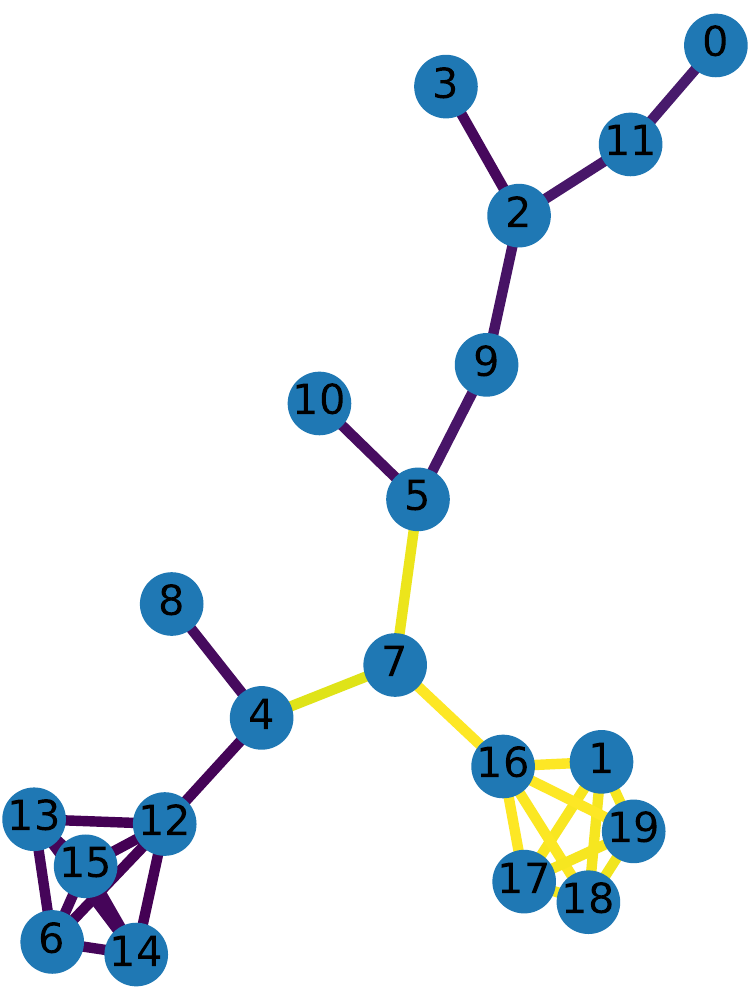}
\caption{GNNExplainer}
\label{fig_gnnexplainer}
\end{subfigure}%
\begin{subfigure}{0.23\textwidth}
\includegraphics[width=0.8\linewidth]{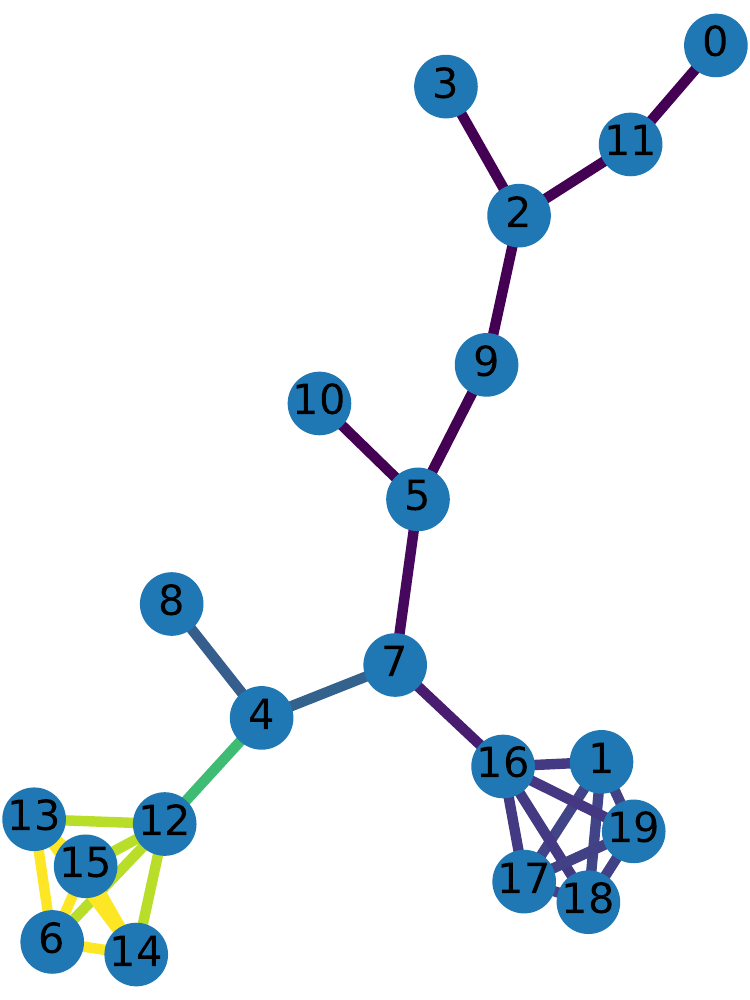}
\caption{Sensitivity}
\label{fig_sensitivity}
\end{subfigure}
\caption{Edge explanations methods for explaining a clique predictor. Edges in cliques should be bright for correct explanations.}
\label{fig_clique}
\end{center}
\vskip -0.2in
\end{figure}

In Figure~\ref{fig_occlusion}, we can see the explanation of occlusion, asking whether the graph contains a clique. Occlusion considers the yellow edge most indicative of the existence of a clique. This edge is not part of a clique. Instead, removing the edge creates a disconnected graph. This graph probably confuses the GNN, which was trained with only connected graphs. This phenomenon can happen whenever we use graphs for explanations that substantially differ from the training data. The explanation of the target class might blend with an adversarial attack against it. Thus we find a misleading (in this case even wrong) explanation. Therefore, we argue to only ever use data consistent with the training distribution for making model explanations. We call this requirement doing \emph{Distribution Compliant Explanation} (DCE).

We propose a novel method CoGE (Contrastive Gnn Explanation) based on contrastive explanation~\cite{dhurandhar2018explanations, dhurandhar2019model}. CoGE aims to find similarities to graphs with the same label and differences to graphs with a different label. Using only existing graphs for explanation, CoGE clearly fulfills DCE. In an experimental evaluation, we show the efficacy of CoGE. Our contributions are:
\begin{itemize}
    \item We motivated the necessity of DCE for explaining GNN predictions.
    \item We present CoGE, a novel method for explaining GNN predictions for graph classification. CoGE uses a contrastive approach and adheres to DCE.
    \item We show the efficacy of our method empirically on existing and synthetic datasets.
\end{itemize}

\section{Related work}
\label{related}

\paragraph{Graph Neural Networks}
Starting with~\citet{scarselli2008graph}, graph neural networks have achieved remarkable results for graph-structured predictions~\citep{zhou2018graph, wu2020comprehensive}. Since then, one family of graph neural networks was developed as generalized convolutional networks~\citep{duvenaud2015convolutional, kipf2017graph} where nodes update their embedding via aggregating their neighborhood~\cite{gilmer2017neural, hamilton2017inductive, battaglia2018relational, xu2019powerful}. Other propagation methods such as attention or skip-connections have also been proposed \citep{velickovic2018graph, xu2018jumping}.

\paragraph{Explainability Methods for Graphs}
GNNExplainer explains predictions by finding an edge mask and feature mask that maximize the mutual information between the prediction and the masked substructure \cite{ying2019gnnexplainer}. Image attribution methods \cite{ancona2017unified} such as  saliency maps and Layer-wise Relevance Propagation have been adapted to graph data structures \cite{baldassarre2019explainability,pope2019explainability, huang2020graphlime}. These methods do not consider the data distribution. For example, an edge mask can ``remove'' edges from the graph and thus, similar to occlusion, lead to misleading explanations.

\paragraph{Adversarial Graph Attacks}
Adversarial attacks on graphs exploit that graph neural networks behave in an undefined way outside of the kind of data they saw during training. Several attack angles have been applied successfully\citep{dai2018adversarial, zugner2018adversarial, zugner2019adversarial, xu2019topology}. At that, \citet{zugner2018adversarial} noted that already small perturbations lead to large changes in the prediction. In this, we see support in our claim that DCE is important for explanations as DCE work contrary to adversarial attacks. Adversarial attacks find and exploit out-of-distribution anomalies in the model, whereas a DCE conforming method aims to explicitly disregard these anomalies and instead searches for genuine patterns.

\section{Method}
\label{method}
\paragraph{Preliminaries}
We consider GNNs operating on undirected graphs $G=(V,E)$ labeled by $y(G)$, with node set $V$ and edge set $E$. Nodes can have attributes in a feature matrix $\mathbf{X}$. We assume a learning model that transforms these initial node features into final embeddings that are aggregated for a graph-level representation. One example of such networks is Message Passing Graph Neural Networks~\citep{gilmer2017neural} that iteratively update node embeddings based on their immediate neighbors.

\paragraph{Explanations for graph classification}
We follow a contrastive approach~\citep{dhurandhar2018explanations, dhurandhar2019model}. Such an approach bases the explanation on other graphs from training, and thus fulfills DCE. In particular, we find the parts of the graph that make this graph distant to graphs with a different label and close to graphs with the same label.

To this end, we need a measure to compare the distance between graphs. In our case, it suffices to define how graphs differ for the classification problem. There exist many graph distance and similarity measures~\citep{sanfeliu1983distance, heimann2018regal, zhang2019deep, wang2019learning} in the literature but they compare principally and offer more than we need. For our explanation purpose, we opt to compare the similarity in the model embedding space. Thus, we can leverage the model information which structures and features are relevant or semantically equivalent for the classification problem. For computing the similarity of two graphs (with respect to the classification problem), we measure a set to set distance of the final node embeddings.

\begin{figure}[ht]
\begin{center}
\def\svgwidth{0.9\columnwidth}
\usetikzlibrary{fadings}
\usetikzlibrary{patterns}
\usetikzlibrary{shadows.blur}
\usetikzlibrary{shapes}

\tikzset{every picture/.style={line width=0.75pt}} 

\begin{tikzpicture}[x=0.75pt,y=0.75pt,yscale=-0.7,xscale=0.7]

\draw [color={rgb, 255:red, 200; green, 200; blue, 200 }  ,draw opacity=1 ]   (175,50) .. controls (214.08,-0.42) and (271.32,27.15) .. (288.31,47.79) ;
\draw [shift={(290,50)}, rotate = 235.05] [fill={rgb, 255:red, 200; green, 200; blue, 200 }  ,fill opacity=1 ][line width=0.08]  [draw opacity=0] (8.93,-4.29) -- (0,0) -- (8.93,4.29) -- cycle    ;
\draw [color={rgb, 255:red, 200; green, 200; blue, 200 }  ,draw opacity=1 ]   (75,50) .. controls (126.19,7.83) and (223.96,18.18) .. (243.64,47.69) ;
\draw [shift={(245,50)}, rotate = 242.89] [fill={rgb, 255:red, 200; green, 200; blue, 200 }  ,fill opacity=1 ][line width=0.08]  [draw opacity=0] (8.93,-4.29) -- (0,0) -- (8.93,4.29) -- cycle    ;
\draw [color={rgb, 255:red, 200; green, 200; blue, 200 }  ,draw opacity=1 ]   (125,80) .. controls (175.21,121.19) and (214.97,107.96) .. (242.88,82.02) ;
\draw [shift={(245,80)}, rotate = 495.78] [fill={rgb, 255:red, 200; green, 200; blue, 200 }  ,fill opacity=1 ][line width=0.08]  [draw opacity=0] (8.93,-4.29) -- (0,0) -- (8.93,4.29) -- cycle    ;
\draw [color={rgb, 255:red, 200; green, 200; blue, 200 }  ,draw opacity=1 ]   (125,80) .. controls (165.59,126.78) and (255.19,151.26) .. (293.84,82.12) ;
\draw [shift={(295,80)}, rotate = 477.91] [fill={rgb, 255:red, 200; green, 200; blue, 200 }  ,fill opacity=1 ][line width=0.08]  [draw opacity=0] (8.93,-4.29) -- (0,0) -- (8.93,4.29) -- cycle    ;
\draw  [fill={rgb, 255:red, 253; green, 248; blue, 80 }  ,fill opacity=1 ] (60,65) .. controls (60,56.72) and (66.72,50) .. (75,50) .. controls (83.28,50) and (90,56.72) .. (90,65) .. controls (90,73.28) and (83.28,80) .. (75,80) .. controls (66.72,80) and (60,73.28) .. (60,65) -- cycle ;
\draw [line width=1.5]    (210,30) -- (210,100) ;
\draw  [fill={rgb, 255:red, 244; green, 159; blue, 79 }  ,fill opacity=1 ] (110,65) .. controls (110,56.72) and (116.72,50) .. (125,50) .. controls (133.28,50) and (140,56.72) .. (140,65) .. controls (140,73.28) and (133.28,80) .. (125,80) .. controls (116.72,80) and (110,73.28) .. (110,65) -- cycle ;
\draw  [fill={rgb, 255:red, 244; green, 159; blue, 79 }  ,fill opacity=1 ] (230,65) .. controls (230,56.72) and (236.72,50) .. (245,50) .. controls (253.28,50) and (260,56.72) .. (260,65) .. controls (260,73.28) and (253.28,80) .. (245,80) .. controls (236.72,80) and (230,73.28) .. (230,65) -- cycle ;
\draw  [fill={rgb, 255:red, 225; green, 225; blue, 50 }  ,fill opacity=1 ] (350,65) .. controls (350,56.72) and (356.72,50) .. (365,50) .. controls (373.28,50) and (380,56.72) .. (380,65) .. controls (380,73.28) and (373.28,80) .. (365,80) .. controls (356.72,80) and (350,73.28) .. (350,65) -- cycle ;
\draw  [fill={rgb, 255:red, 238; green, 96; blue, 98 }  ,fill opacity=1 ] (160,65) .. controls (160,56.72) and (166.72,50) .. (175,50) .. controls (183.28,50) and (190,56.72) .. (190,65) .. controls (190,73.28) and (183.28,80) .. (175,80) .. controls (166.72,80) and (160,73.28) .. (160,65) -- cycle ;
\draw  [fill={rgb, 255:red, 226; green, 66; blue, 45 }  ,fill opacity=1 ] (280,65) .. controls (280,56.72) and (286.72,50) .. (295,50) .. controls (303.28,50) and (310,56.72) .. (310,65) .. controls (310,73.28) and (303.28,80) .. (295,80) .. controls (286.72,80) and (280,73.28) .. (280,65) -- cycle ;
\draw  [fill={rgb, 255:red, 253; green, 247; blue, 52 }  ,fill opacity=1 ] (400,65) .. controls (400,56.72) and (406.72,50) .. (415,50) .. controls (423.28,50) and (430,56.72) .. (430,65) .. controls (430,73.28) and (423.28,80) .. (415,80) .. controls (406.72,80) and (400,73.28) .. (400,65) -- cycle ;
\draw [line width=1.5]    (330,30) -- (330,100) ;

\draw (118,-0.5) node [anchor=north west][inner sep=0.75pt]   [align=left] {$\displaystyle G$};
\draw (261,-1) node [anchor=north west][inner sep=0.75pt]   [align=left] {$\displaystyle G^{\not{\approx }}$};
\draw (381,-1) node [anchor=north west][inner sep=0.75pt]   [align=left] {$\displaystyle G^{\approx }$};
\draw (69.5,55.5) node [anchor=north west][inner sep=0.75pt]   [align=left] {1};
\draw (120,55.5) node [anchor=north west][inner sep=0.75pt]   [align=left] {2};
\draw (169.5,55.5) node [anchor=north west][inner sep=0.75pt]   [align=left] {3};
\draw (240,55.5) node [anchor=north west][inner sep=0.75pt]   [align=left] {4};
\draw (290,55.5) node [anchor=north west][inner sep=0.75pt]   [align=left] {5};
\draw (359.5,55.5) node [anchor=north west][inner sep=0.75pt]   [align=left] {6};
\draw (409.5,55.5) node [anchor=north west][inner sep=0.75pt]   [align=left] {7};
\draw (67,80) node [anchor=north west][inner sep=0.75pt]   [align=left] {$\displaystyle \nicefrac{1}{3}$};
\draw (117,80) node [anchor=north west][inner sep=0.75pt]   [align=left] {$\displaystyle \nicefrac{1}{3}$};
\draw (167,80) node [anchor=north west][inner sep=0.75pt]   [align=left] {$\displaystyle \nicefrac{1}{3}$};
\draw (237,80) node [anchor=north west][inner sep=0.75pt]   [align=left] {$\displaystyle \nicefrac{1}{2}$};
\draw (287,80) node [anchor=north west][inner sep=0.75pt]   [align=left] {$\displaystyle \nicefrac{1}{2}$};
\draw (357,80) node [anchor=north west][inner sep=0.75pt]   [align=left] {$\displaystyle \nicefrac{1}{2}$};
\draw (407,80) node [anchor=north west][inner sep=0.75pt]   [align=left] {$\displaystyle \nicefrac{1}{2}$};
\draw (151,3) node [anchor=north west][inner sep=0.75pt]   [align=left] {$\displaystyle \nicefrac{1}{3}$};
\draw (215,2) node [anchor=north west][inner sep=0.75pt]   [align=left] {$\displaystyle \nicefrac{1}{3}$};
\draw (195,103) node [anchor=north west][inner sep=0.75pt]   [align=left] {$\displaystyle \nicefrac{1}{6}$};
\draw (221,124) node [anchor=north west][inner sep=0.75pt]   [align=left] {$\displaystyle \nicefrac{1}{6}$};

\end{tikzpicture}
\vskip -0.2in
\caption{Example for optimal transport (OT).}
\label{fig_transport}
\end{center}
\end{figure}
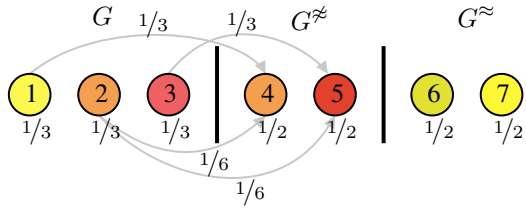

In particular, we measure using the Optimal Transport (OT) distance \citep{nikolentzos2017matching, fey2020deep}. Figure~\ref{fig_transport} shows an example of computing OT between the left and middle graphs. All nodes have a weight (such that the weights of all nodes per graph sum to $1$. Now, every node from the source needs to transport its weight to one or more target nodes, whose weight denotes their maximum capacity. The cost of one transport is the transport weight times the distance between the node embeddings --- we use $L_2$ distance for this. Optimal transport finds the globally optimal soft assignment, even if this involves suboptimal choices for some nodes. For example in Figure~\ref{fig_transport}, node $2$ does not move all its weight to node $4$, even though they are the same and the cost would be $0$. Thus, optimal transport allows us to compare embeddings of two graphs on node granularity. We compare this to a graph-level metric in the experiments where we measure the $L_2$ distance between the average node embeddings.

But we can also change the initial node weights to something different than uniform. If we want to change the source weights as to \emph{minimize} optimal transport, the nodes that do not have counterparts in the target graph receive a low weight. If we want to change source weights as to \emph{maximize} optimal transport, the nodes with counterparts receive a low weight. These two observations give us a framework to find explaining graph parts. Jointly maximizing OT to graphs with the same label and minimizing OT to graphs with a different label finds us the explanation nodes. These are the nodes with a low weight. Formally, we want to find weights $w_{opt}$ for the nodes of graph G such that
\begin{equation}\label{eq_loss}
w_{opt}(G) = \argmin_w \mathcal{L}^{\not\approx}_w(G)- \mathcal{L}_w^\approx(G) + \mathcal{L}_w^=(G)
\end{equation}
The first loss term captures the average distance to the $k$ most similar graphs with a different label $\mathbb{G}_k^{\not\approx}$. The second loss term captures the average distance to the $k$ most similar graphs with the same label $\mathbb{G}_k^{\approx}$. We compute the similarity to choose the graphs with uniformly weighted optimal transport. The hyperparameter $k$ determines how many graphs to compare against. The third term compares the distance of the $w$-weighted graph G to its uniformly-weighted version. This term acts as regularization. It penalizes any deviation from uniform weights, thus biasing $w$ to only make few modifications with substantial benefits. Let $d(Z_G, w, Z_H)$ be the optimal transport distance between two sets of embeddings from graphs $G$ and $H$, where we weight $Z_G$ according to $w$ and $Z_H$ with uniform weights. Formally the losses are then defined as:
\begin{align}
\mathcal{L}_W^{\not\approx}(G) &= \frac{1}{k} \sum_{H \in \mathbb{G}_k^{\not\approx}} d_W(Z_G, Z_H)\nonumber\\
\mathcal{L}_W^\approx(G) &= \frac{1}{k} \sum_{H \in \mathbb{G}_k^{\approx}} d_W(Z_G, Z_H)\nonumber\\
\mathcal{L}_W^=(G) &= d_W(Z_G, Z_G)\nonumber
\end{align}

\section{Experiments}
\subsection{CoGE Implementation}
For CoGE, we set the number $k$ of graphs to contrast again to $10$. Our framework uses the OT as provided by the GeomLoss library~\citep{geomloss}. For optimization of Equation~\eqref{eq_loss}, we use gradient descent using the Adam optimizer~\citep{adam} with a learning rate of $0.1$, except for REDDIT, where we use $0.01$\footnote{Code available at \url{https://github.com/lukasjf/contrastive-gnn-explanation}}.

\subsection{Qualitative Analysis}
We analyze our explanations on two well known real-world datasets for graph classification: MUTAG~\citep{mutag} and REDDIT-BINARY~\citep{reddit}. MUTAG labels $4337$ chemical molecules for their mutagenic effect. REDDIT-BINARY classifies $2000$ Reddit threads whether they are of type Q\&A or discussion. We train Graph Isomorphism networks~\citep{xu2019powerful}, which achieve state-of-the-art performance on both datasets and analyze the trained models. 

\begin{figure*}[ht]
\vskip 0.2in
\begin{center}
\def\svgwidth{\textwidth}
\centerline{\input{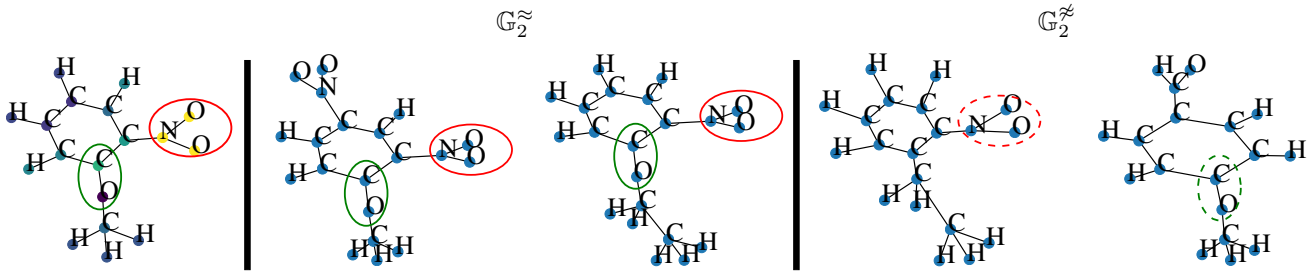}}
\caption{MUTAG Explanation showing important substructures in molecules. Left: Original graph, Middle: Similar graphs with the same label, Right: Similar graphs with a different label. Brighter nodes are more important in the explanation.}
\label{fig_mutag}
\end{center}
\vskip -0.25in
\end{figure*}

In MUTAG (see Figure~\ref{fig_mutag}), our method identifies the $NO_2$ structure (red circle) as being primarily important, which is a known mutagenic part~\citep{mutag}. However, it is also present in a few non-mutagenic graphs. As a second explanation, our method identifies the $C$ next to an $O$ close to this component (green circle). The combination of structures is only present in the mutagenic examples.

\begin{figure}[ht]
\vskip 0.2in
\begin{center}
\centerline{\includegraphics[width=0.7\columnwidth]{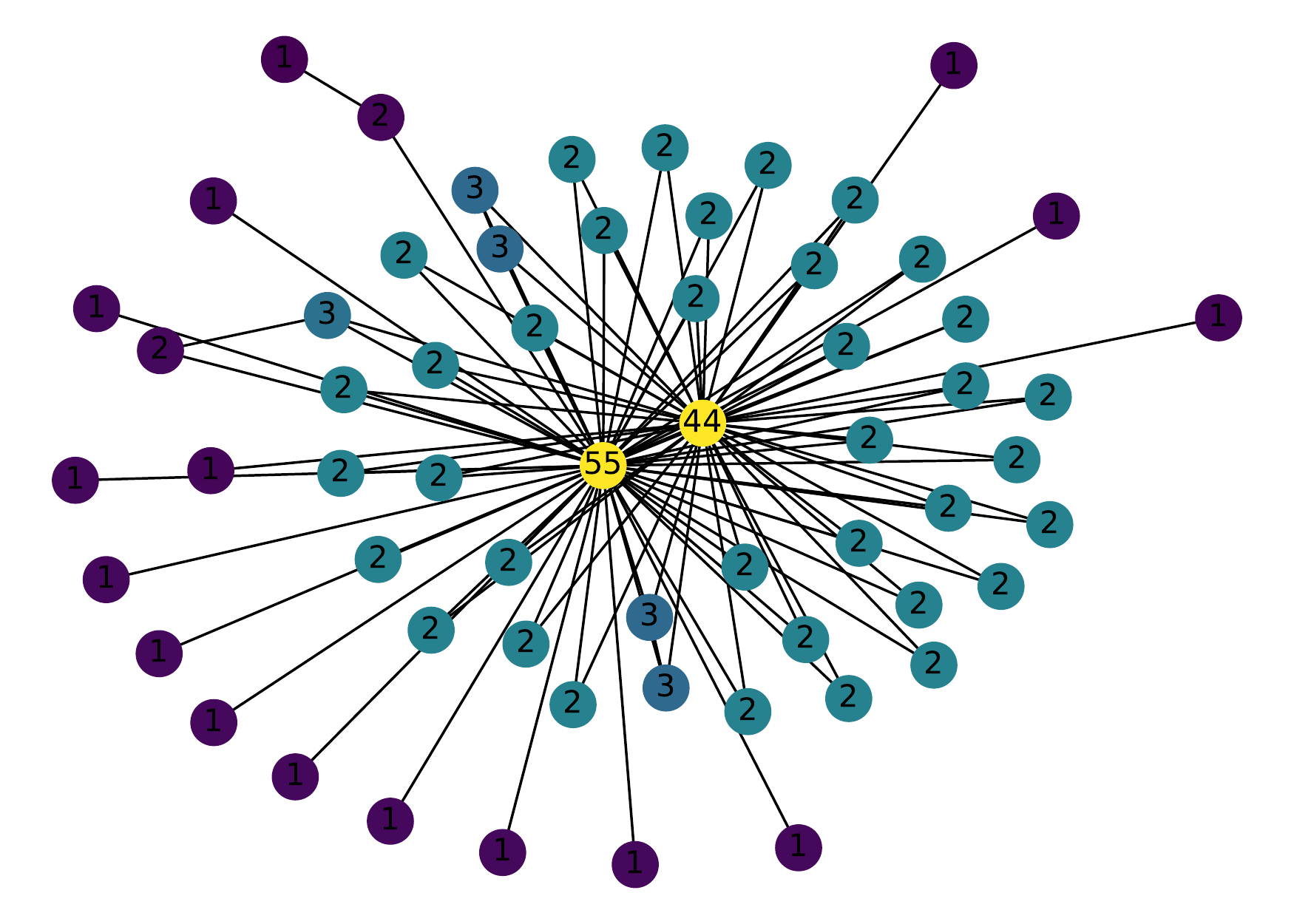}}
\caption{Example Q\&A explanation. Numbers depict the node degree. Nodes connected to both central nodes are more important.}
\label{fig_reddit}
\end{center}
\vskip -0.25in
\end{figure}

In REDDIT-BINARY, our method considers the central and adjacent nodes to be important for classifying this graph as Q\&A (see Figure~\ref{fig_reddit}). They are indeed important since a Q\&A threads consist of few experts (high degree nodes) and most users ask them questions and getting replies. In contrast, discussions typically have only one central node and the graph has a tree-like structure with higher depth.

\subsection{Quantitative Analysis}
\label{quant}
\paragraph{Dataset}
We present \emph{CYCLIQ} (cycles and cliques), a new dataset for explaining graph classifications similar to the node classification dataset in \citet{ying2019gnnexplainer}. The CYCLIQ dataset is a binary classification problem. CYCLIQ's primary building blocks are random trees, to which we append either cycles or cliques. The target labels store if the graph has cycles or cliques. Nodes have an initial feature vector of size 10, initialized to all ones. The correct explanations for this dataset are the edges in a clique or cycle structure. For evaluation, we count the number $x$ edges being part of a cycle or a clique. The explanation accuracy of a method is the ratio of how many edges in cycle or cliques are in the method's $x$ most important edges.

\paragraph{Experiment Setup}
We run all explanations on a GCN~\citep{kipf2017graph} with $5$ layers. We use a constant embedding size of $20$ across all layers. There are no edge features. We create a total of $2000$ graphs, using an $80/20$ train-test split. The model understands the task well, reaching $99\%$ test accuracy. We compare our method against random guessing, a node-based occlusion method (removing adjacent edges), sensitivity analysis, and GNNExplainer~\cite{ying2019gnnexplainer} using the provided implementation\footnote{\url{https://github.com/RexYing/gnn-model-explainer}}. To allow comparison with GNNExplainer we focus on edge importances by summing the importance of adjacent nodes for node-based explanation methods.

\paragraph{Results}
We report the explanation accuracies across the $1600$ training graphs in table~\ref{tab_explanation-accs}. Generally, it seems easier to explain cliques than it is to explain cycles. We note that CoGE produces the best results. For both classes it outperforms the other methods by at least $10\%$, reaching almost perfect accuracy for cliques. Visual example explanations for all methods (but random) are in Figure~\ref{fig_clique}.

\begin{table}[t]
\caption{Explanation accuracies on the CYCLIQ dataset.}
\label{tab_explanation-accs}
\vskip 0.15in
\begin{center}
\begin{small}
\begin{tabular}{lccc}
\toprule
Method & Cycle Acc. & Clique Acc. & Avg. Acc.\\
\midrule
Random   & 0.41 $\pm$ 0.17 & 0.58 $\pm$ 0.13 & 0.49  \\
Occlusion    & 0.39 $\pm$ 0.23 & 0.86 $\pm$ 0.16 & 0.62  \\
Sensitivity    & 0.36 $\pm$ 0.2 & 0.87 $\pm$ 0.12 & 0.61  \\
GNNExplainer    & 0.43 $\pm$ 0.18 & 0.73 $\pm$ 0.14 & 0.58  \\
\textbf{CoGE}    & \textbf{0.78 $\pm$ 0.18} & \textbf{0.99 $\pm$ 0.02} & \textbf{0.88} \\
\bottomrule
\end{tabular}
\end{small}
\end{center}
\vskip -0.1in
\end{table}

\paragraph{Ablation Study}
Next, we study the importance of each loss term from Equation~\eqref{eq_loss}, reusing above CYCLIQ setting. Table~\ref{table_ablation} shows that $\mathcal{L}_W^{\not\approx}$ captures most explanation, but $\mathcal{L}_W^{\approx}$ and $\mathcal{L}_W^=$ help improve further. Additionally, we try replacing OT distance with the euclidean distance on the weighted average on the node embeddings ($\mathcal{L}$ and Average). This leads to worse accuracy while still outperforming baselines.

\begin{table}[t]
\caption{Explanation accuracies on the CYCLIQ dataset for ablations of the loss and the distance measure.}
\label{table_ablation}
\vskip 0.15in
\begin{center}
\begin{small}
\begin{tabular}{lccc}
\toprule
Loss & Cycle Acc. & Clique Acc. & Avg. Acc.\\
\midrule
$-\mathcal{L}_W^{\approx}$ & 0.63 $\pm$ 0.25 & 0.6 $\pm$ 0.35 & 0.62  \\
$-\mathcal{L}_W^{\approx}  + \mathcal{L}_W^{=}$ & 0.62 $\pm$ 0.23 & 0.61 $\pm$ 0.27 & 0.61  \\
$\mathcal{L}_W^{\not{\approx}}$    & 0.65 $\pm$ 0.23 & 0.99 $\pm$ 0.02 & 0.81 \\
$\mathcal{L}_W^{\not{\approx}} + \mathcal{L}_W^{=}$ & 0.66 $\pm$ 0.24 & 0.99 $\pm$ 0.02 & 0.82  \\
$\mathcal{L}_W^{\not{\approx}} - \mathcal{L}_W^{\approx}$ & 0.7 $\pm$ 0.2 & 0.99 $\pm$ 0.02 & 0.84 \\
\midrule
$\mathcal{L}_W$ and Average & 0.45 $\pm$ 0.24 & 0.99 $\pm$ 0.02 & 0.71 \\
\midrule
$\mathcal{L}_W$ and OT  & 0.78 $\pm$ 0.18 & 0.99 $\pm$ 0.02 & 0.88 \\
\bottomrule
\end{tabular}
\end{small}
\end{center}
\vskip -0.1in
\end{table}

\section{Conclusion}
In this work, we discuss the particularities of explaining GNN predictions. In graphs, structure is important, as only slight modifications can lead to graphs out of the known data distribution. Explanations start to blend with adversarial attacks. Therefore, we argue that explanation methods should stay with the training data distribution and produce \emph{Distribution Compliant Explanation} (DCE). We propose a novel explanation method, CoGE, for graph classification that adheres to DCE. Experimental results verify its effectiveness and its robustness to parameter choices. Fur future work, we aim at extending our findings to node classification and better understanding the connection between explanation and adversarial attacks.

\bibliography{example_paper}

\begin{thebibliography}{33}
\providecommand{\natexlab}[1]{#1}
\providecommand{\url}[1]{\texttt{#1}}
\expandafter\ifx\csname urlstyle\endcsname\relax
  \providecommand{\doi}[1]{doi: #1}\else
  \providecommand{\doi}{doi: \begingroup \urlstyle{rm}\Url}\fi

\bibitem[Ancona et~al.(2017)Ancona, Ceolini, \"Oztireli, and
  Gross]{ancona2017unified}
Ancona, M., Ceolini, E., \"Oztireli, C., and Gross, M.
\newblock A unified view of gradient-based attribution methods for deep neural
  networks.
\newblock In \emph{{NIPS} Workshop on Interpreting, Explaining and Visualizing
  Deep Learning, Long Beach, USA}, December 2017.

\bibitem[Baldassarre \& Azizpour(2019)Baldassarre and
  Azizpour]{baldassarre2019explainability}
Baldassarre, F. and Azizpour, H.
\newblock Explainability techniques for graph convolutional networks.
\newblock \emph{arXiv preprint arXiv:1905.13686}, 2019.

\bibitem[Battaglia et~al.(2018)Battaglia, Hamrick, Bapst, Sanchez-Gonzalez,
  Zambaldi, Malinowski, Tacchetti, Raposo, Santoro, Faulkner,
  et~al.]{battaglia2018relational}
Battaglia, P.~W., Hamrick, J.~B., Bapst, V., Sanchez-Gonzalez, A., Zambaldi,
  V., Malinowski, M., Tacchetti, A., Raposo, D., Santoro, A., Faulkner, R.,
  et~al.
\newblock Relational inductive biases, deep learning, and graph networks.
\newblock \emph{arXiv preprint arXiv:1806.01261}, 2018.

\bibitem[Dai et~al.(2018)Dai, Li, Tian, Huang, Wang, Zhu, and
  Song]{dai2018adversarial}
Dai, H., Li, H., Tian, T., Huang, X., Wang, L., Zhu, J., and Song, L.
\newblock Adversarial attack on graph structured data.
\newblock In \emph{Proceedings of the International Conference on Machine
  Learning (ICML), Stockholm, Sweden}, July 2018.

\bibitem[Debnath et~al.(1991)Debnath, Lopez~de Compadre, Debnath, Shusterman,
  and Hansch]{mutag}
Debnath, A.~K., Lopez~de Compadre, R.~L., Debnath, G., Shusterman, A.~J., and
  Hansch, C.
\newblock Structure-activity relationship of mutagenic aromatic and
  heteroaromatic nitro compounds. correlation with molecular orbital energies
  and hydrophobicity.
\newblock \emph{Journal of medicinal chemistry}, 1991.

\bibitem[Dhurandhar et~al.(2018)Dhurandhar, Chen, Luss, Tu, Ting, Shanmugam,
  and Das]{dhurandhar2018explanations}
Dhurandhar, A., Chen, P.-Y., Luss, R., Tu, C.-C., Ting, P., Shanmugam, K., and
  Das, P.
\newblock Explanations based on the missing: Towards contrastive explanations
  with pertinent negatives.
\newblock In \emph{Advances in Neural Information Processing Systems}, 2018.

\bibitem[Dhurandhar et~al.(2019)Dhurandhar, Pedapati, Balakrishnan, Chen,
  Shanmugam, and Puri]{dhurandhar2019model}
Dhurandhar, A., Pedapati, T., Balakrishnan, A., Chen, P.-Y., Shanmugam, K., and
  Puri, R.
\newblock Model agnostic contrastive explanations for structured data.
\newblock \emph{arXiv preprint arXiv:1906.00117}, 2019.

\bibitem[Duvenaud et~al.(2015)Duvenaud, Maclaurin, Iparraguirre, Bombarell,
  Hirzel, Aspuru-Guzik, and Adams]{duvenaud2015convolutional}
Duvenaud, D.~K., Maclaurin, D., Iparraguirre, J., Bombarell, R., Hirzel, T.,
  Aspuru-Guzik, A., and Adams, R.~P.
\newblock Convolutional networks on graphs for learning molecular fingerprints.
\newblock In \emph{Advances in Neural Information Processing Systems}, 2015.

\bibitem[Fey et~al.(2020)Fey, Lenssen, Morris, Masci, and Kriege]{fey2020deep}
Fey, M., Lenssen, J.~E., Morris, C., Masci, J., and Kriege, N.~M.
\newblock Deep graph matching consensus.
\newblock \emph{arXiv preprint arXiv:2001.09621}, 2020.

\bibitem[Feydy et~al.(2019)Feydy, S{\'e}journ{\'e}, Vialard, Amari, Trouve, and
  Peyr{\'e}]{geomloss}
Feydy, J., S{\'e}journ{\'e}, T., Vialard, F.-X., Amari, S.-i., Trouve, A., and
  Peyr{\'e}, G.
\newblock Interpolating between optimal transport and mmd using sinkhorn
  divergences.
\newblock In \emph{Proceedings of the International Conference on Artificial
  Intelligence and Statistics (AISTATS), Naha, Japan}, April 2019.

\bibitem[Gilmer et~al.(2017)Gilmer, Schoenholz, Riley, Vinyals, and
  Dahl]{gilmer2017neural}
Gilmer, J., Schoenholz, S.~S., Riley, P.~F., Vinyals, O., and Dahl, G.~E.
\newblock Neural message passing for quantum chemistry.
\newblock In \emph{Proceedings of the International Conference on Machine
  Learning (ICML), Sydney, Australia}, August 2017.

\bibitem[Hamilton et~al.(2017)Hamilton, Ying, and
  Leskovec]{hamilton2017inductive}
Hamilton, W., Ying, Z., and Leskovec, J.
\newblock Inductive representation learning on large graphs.
\newblock In \emph{Advances in Neural Information Processing Systems}, 2017.

\bibitem[Heimann et~al.(2018)Heimann, Shen, Safavi, and
  Koutra]{heimann2018regal}
Heimann, M., Shen, H., Safavi, T., and Koutra, D.
\newblock Regal: Representation learning-based graph alignment.
\newblock In \emph{Proceedings of the ACM International Conference on
  Information and Knowledge Management (CIKM), Turin, Italy}, October 2018.

\bibitem[Huang et~al.(2020)Huang, Yamada, Tian, Singh, Yin, and
  Chang]{huang2020graphlime}
Huang, Q., Yamada, M., Tian, Y., Singh, D., Yin, D., and Chang, Y.
\newblock Graphlime: Local interpretable model explanations for graph neural
  networks.
\newblock \emph{arXiv preprint arXiv:2001.06216}, 2020.

\bibitem[Kingma \& Ba(2014)Kingma and Ba]{adam}
Kingma, D.~P. and Ba, J.
\newblock Adam: A method for stochastic optimization.
\newblock \emph{arXiv preprint arXiv:1412.6980}, 2014.

\bibitem[Kipf \& Welling(2017)Kipf and Welling]{kipf2017graph}
Kipf, T.~N. and Welling, M.
\newblock Semi-supervised classification with graph convolutional networks.
\newblock In \emph{Proceedings of the International Conference on Learning
  Representations (ICLR), Toulon, France Toulon, France, April 24-26, 2017,
  Conference Track Proceedings}, April 2017.

\bibitem[Nikolentzos et~al.(2017)Nikolentzos, Meladianos, and
  Vazirgiannis]{nikolentzos2017matching}
Nikolentzos, G., Meladianos, P., and Vazirgiannis, M.
\newblock Matching node embeddings for graph similarity.
\newblock In \emph{Proceedings of the Conference on Artificial Intelligence
  (AAAI), San Francisco, USA}, February 2017.

\bibitem[Pope et~al.(2019)Pope, Kolouri, Rostami, Martin, and
  Hoffmann]{pope2019explainability}
Pope, P.~E., Kolouri, S., Rostami, M., Martin, C.~E., and Hoffmann, H.
\newblock Explainability methods for graph convolutional neural networks.
\newblock In \emph{Proceedings of the IEEE Conference on Computer Vision and
  Pattern Recognition (CVPR), Long Beach, USA}, June 2019.

\bibitem[Sanfeliu \& Fu(1983)Sanfeliu and Fu]{sanfeliu1983distance}
Sanfeliu, A. and Fu, K.-S.
\newblock A distance measure between attributed relational graphs for pattern
  recognition.
\newblock \emph{IEEE transactions on systems, man, and cybernetics}, 1983.

\bibitem[Scarselli et~al.(2008)Scarselli, Gori, Tsoi, Hagenbuchner, and
  Monfardini]{scarselli2008graph}
Scarselli, F., Gori, M., Tsoi, A.~C., Hagenbuchner, M., and Monfardini, G.
\newblock The graph neural network model.
\newblock \emph{IEEE Transactions on Neural Networks}, 2008.

\bibitem[Veli{\v{c}}kovi{\'{c}} et~al.(2018)Veli{\v{c}}kovi{\'{c}}, Cucurull,
  Casanova, Romero, Li{\`{o}}, and Bengio]{velickovic2018graph}
Veli{\v{c}}kovi{\'{c}}, P., Cucurull, G., Casanova, A., Romero, A., Li{\`{o}},
  P., and Bengio, Y.
\newblock {Graph Attention Networks}.
\newblock \emph{Proceedings of the International Conference on Learning
  Representations (ICLR), Vancouver, Canada}, May 2018.

\bibitem[Wang et~al.(2019)Wang, Yan, and Yang]{wang2019learning}
Wang, R., Yan, J., and Yang, X.
\newblock Learning combinatorial embedding networks for deep graph matching.
\newblock In \emph{Proceedings of the IEEE International Conference on Computer
  Vision (CVPR), Long Beach, USA}, June 2019.

\bibitem[{Wu} et~al.(2020){Wu}, {Pan}, {Chen}, {Long}, {Zhang}, and
  {Yu}]{wu2020comprehensive}
{Wu}, Z., {Pan}, S., {Chen}, F., {Long}, G., {Zhang}, C., and {Yu}, P.~S.
\newblock A comprehensive survey on graph neural networks.
\newblock \emph{IEEE Transactions on Neural Networks and Learning Systems},
  pp.\  1--21, 2020.

\bibitem[Xu et~al.(2018)Xu, Li, Tian, Sonobe, Kawarabayashi, and
  Jegelka]{xu2018jumping}
Xu, K., Li, C., Tian, Y., Sonobe, T., Kawarabayashi, K., and Jegelka, S.
\newblock Representation learning on graphs with jumping knowledge networks.
\newblock In \emph{Proceedings of the International Conference on Machine
  Learning (ICML), Stockholm, Sweden}, July 2018.

\bibitem[Xu et~al.(2019{\natexlab{a}})Xu, Chen, Liu, Chen, Weng, Hong, and
  Lin]{xu2019topology}
Xu, K., Chen, H., Liu, S., Chen, P.-Y., Weng, T.-W., Hong, M., and Lin, X.
\newblock Topology attack and defense for graph neural networks: An
  optimization perspective.
\newblock In \emph{Proceedings of the International Joint Conference on
  Artificial Intelligence (IJCAI), Macao, China}, July 2019{\natexlab{a}}.

\bibitem[Xu et~al.(2019{\natexlab{b}})Xu, Hu, Leskovec, and
  Jegelka]{xu2019powerful}
Xu, K., Hu, W., Leskovec, J., and Jegelka, S.
\newblock How powerful are graph neural networks?
\newblock In \emph{Proceedings of the International Conference on Learning
  Representations (ICLR) New Orleans, USA}, May 2019{\natexlab{b}}.

\bibitem[Yanardag \& Vishwanathan(2015)Yanardag and Vishwanathan]{reddit}
Yanardag, P. and Vishwanathan, S.
\newblock Deep graph kernels.
\newblock In \emph{Proceedings of the ACM SIGKDD International Conference on
  Knowledge Discovery and Data Mining (KDD), Sydney, Australia}, August 2015.

\bibitem[Ying et~al.(2019)Ying, Bourgeois, You, Zitnik, and
  Leskovec]{ying2019gnnexplainer}
Ying, Z., Bourgeois, D., You, J., Zitnik, M., and Leskovec, J.
\newblock Gnnexplainer: Generating explanations for graph neural networks.
\newblock In \emph{Advances in Neural Information Processing Systems}, 2019.

\bibitem[Zeiler \& Fergus(2014)Zeiler and Fergus]{zeiler2014visualizing}
Zeiler, M.~D. and Fergus, R.
\newblock Visualizing and understanding convolutional networks.
\newblock In \emph{Proceedings of the European Conference on Computer Vision
  (ECCV), Zurich, Switzerland}, September 2014.

\bibitem[Zhang \& Lee(2019)Zhang and Lee]{zhang2019deep}
Zhang, Z. and Lee, W.~S.
\newblock Deep graphical feature learning for the feature matching problem.
\newblock In \emph{Proceedings of the IEEE International Conference on Computer
  Vision (CVPR), Long Beach, USA}, June 2019.

\bibitem[Zhou et~al.(2018)Zhou, Cui, Zhang, Yang, Liu, Wang, Li, and
  Sun]{zhou2018graph}
Zhou, J., Cui, G., Zhang, Z., Yang, C., Liu, Z., Wang, L., Li, C., and Sun, M.
\newblock Graph neural networks: A review of methods and applications.
\newblock \emph{arXiv preprint arXiv:1812.08434}, 2018.

\bibitem[Z{\"{u}}gner \& G{\"{u}}nnemann(2019)Z{\"{u}}gner and
  G{\"{u}}nnemann]{zugner2019adversarial}
Z{\"{u}}gner, D. and G{\"{u}}nnemann, S.
\newblock Adversarial attacks on graph neural networks via meta learning.
\newblock In \emph{Proceedings of the International Conference on Learning
  Representations (ICLR), New Orleans, USA}, May 2019.

\bibitem[Z{\"u}gner et~al.(2018)Z{\"u}gner, Akbarnejad, and
  G{\"u}nnemann]{zugner2018adversarial}
Z{\"u}gner, D., Akbarnejad, A., and G{\"u}nnemann, S.
\newblock Adversarial attacks on neural networks for graph data.
\newblock In \emph{Proceedings of the ACM SIGKDD International Conference on
  Knowledge Discovery and Data Mining (KDD), London, UK}, August 2018.

\end{thebibliography}
\bibliographystyle{icml2020}

\end{document}